\documentclass[conference]{IEEEtran}
\IEEEoverridecommandlockouts
\usepackage{cite}
\usepackage{amsmath,amssymb,amsfonts}
\usepackage{algorithm}
\usepackage{algpseudocode}
\usepackage{graphicx}
\usepackage{textcomp}
\usepackage{xcolor}
\usepackage{bm}
\usepackage{subcaption}
\usepackage{stfloats} 

\def\approxpropto{%
  \def\p{%
    \setbox0=\vbox{\hbox{$\propto$}}%
    \ht0=0.6ex \box0 }%
  \def\s{%
    \vbox{\hbox{$\sim$}}%
  }%
  \mathrel{\raisebox{0.7ex}{%
      \mbox{$\underset{\s}{\p}$}%
    }}%
}

\def\BibTeX{{\rm B\kern-.05em{\sc i\kern-.025em b}\kern-.08em
    T\kern-.1667em\lower.7ex\hbox{E}\kern-.125emX}}
\begin{document}

\title{Exploiting Gradients in Bayesian Inference of Expensive Simulators}

\author{\IEEEauthorblockN{1\textsuperscript{st} Šimon Soldát\textsuperscript{1,2},
2\textsuperscript{nd} Václav Šmídl\textsuperscript{1}}
\IEEEauthorblockA{
\textsuperscript{1}Department of Computer Science, Faculty of Electrical Engineering, Czech Technical University in Prague, Czech Republic \\
\textsuperscript{2}Institute of Plasma Physics of the Czech Academy of Sciences, Prague, Czech Republic \\
soldasim@fel.cvut.cz, smidlva1@fel.cvut.cz}
\thanks{\copyright{} 2026 IEEE. \v{S}. Sold\'{a}t and V. \v{S}m\'{i}dl, ``Exploiting Gradients in Bayesian Inference of Expensive Simulators,'' ICMLT 2026, doi: 10.1109/ICMLT69916.2026.11689621}
}

\maketitle

\begin{abstract}
Simulators based on differential equations are ubiquitous in science and engineering. They are often used in simulation-based inference to evaluate the posterior distribution of the input parameters based on real-world observations of the simulator outputs. However, inference becomes challenging when individual simulator evaluations are computationally expensive. In such cases, a Bayesian optimization-based active learning approach with Gaussian process surrogate models has been used to maximize the information obtained from a limited simulation budget. Recently, gradients of simulator outputs with respect to input parameters have become increasingly available, yet they are rarely exploited for inference. Even though we only need to learn the simulator input-output relationship, gradient information can provide an additional valuable signal to guide the active learning procedure. This is of particular interest in the case of expensive simulators, when sample efficiency is crucial.

In this paper, we demonstrate how incorporating gradient information into the Gaussian process surrogate accelerates Bayesian optimization-based inference under a limited simulation budget. Our results show significant improvement in convergence speed from using gradient information. For reverse-mode differentiation, the inference efficiency gains are maintained when accounting for the additional computational cost. In contrast, for forward-mode differentiation, the inference speed-up does not outweigh the computational costs. These results indicate that gradient-enhanced surrogates are beneficial primarily in problems where the number of parameters exceeds the output dimensionality, where reverse-mode differentiation is efficient.
\end{abstract}

\begin{IEEEkeywords}
simulation-based inference, Bayesian inference, Bayesian optimization, Gaussian process, adjoint gradients
\end{IEEEkeywords}

\section{Introduction}

\begin{figure*}[t]
\centering

\begin{subfigure}{0.24\textwidth}
    \centering
    \includegraphics[width=\textwidth]{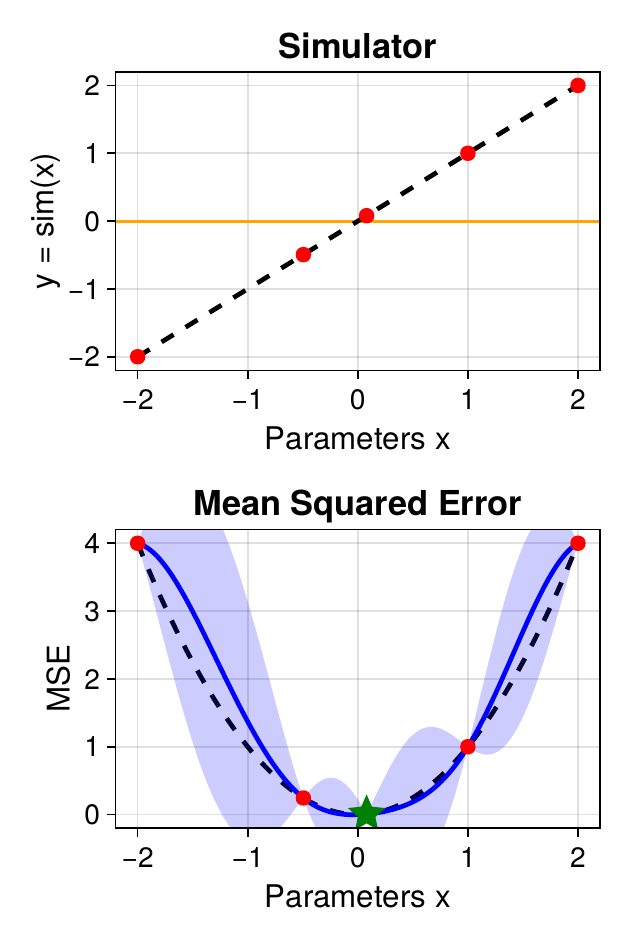}
    \caption{\small Iter. 5}
\end{subfigure}
\begin{subfigure}{0.24\textwidth}
    \centering
    \includegraphics[width=\textwidth]{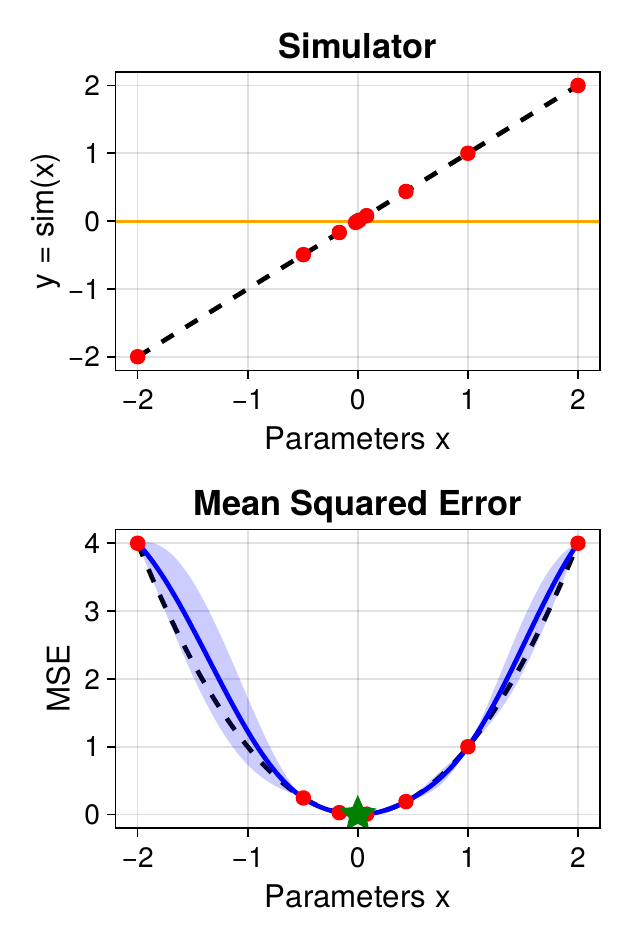}
    \caption{\small Iter. 10}
\end{subfigure}
\begin{subfigure}{0.24\textwidth}
    \centering
    \includegraphics[width=\textwidth]{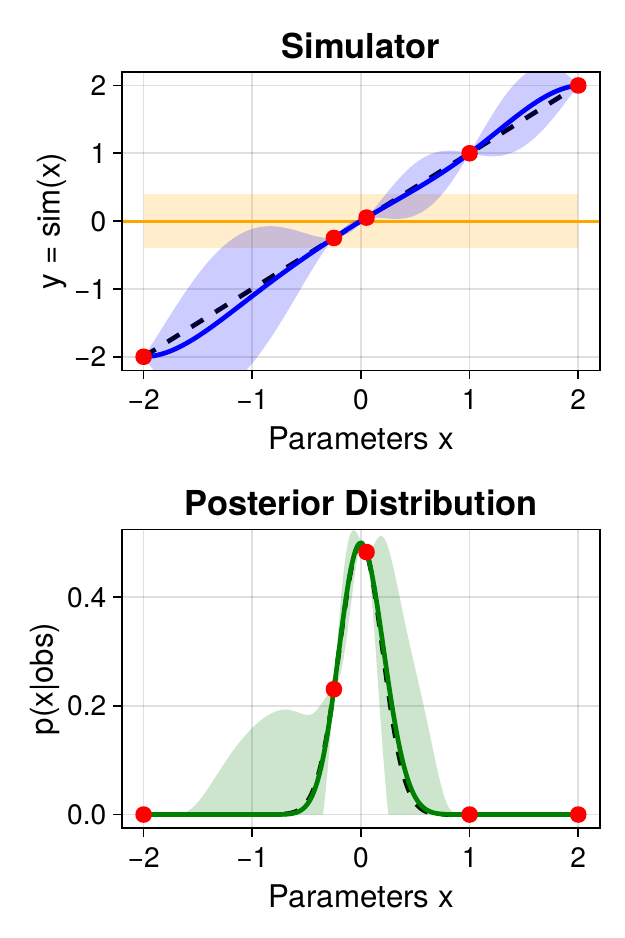}
    \caption{\small Iter. 5}
\end{subfigure}
\begin{subfigure}{0.24\textwidth}
    \centering
    \includegraphics[width=\textwidth]{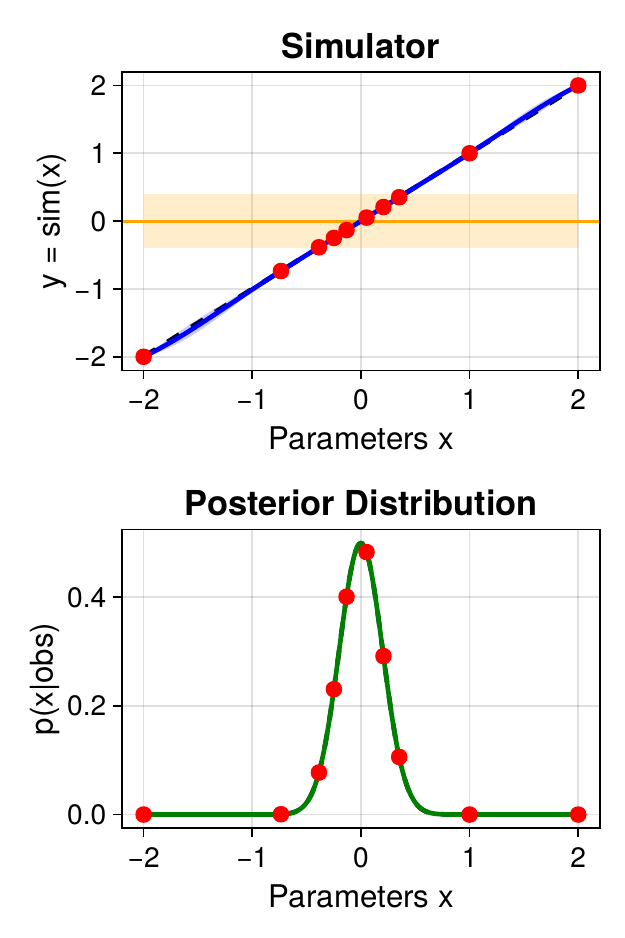}
    \caption{\small Iter. 10}
\end{subfigure}

\begin{subfigure}{0.35\textwidth}
    \centering
    \includegraphics[width=\textwidth]{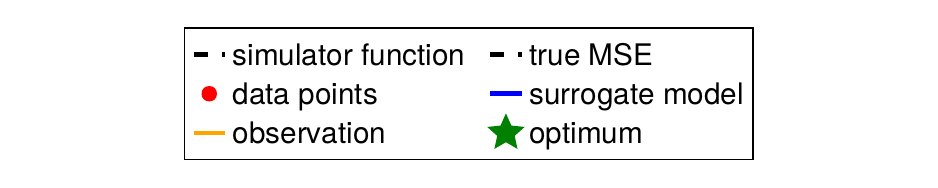}
\end{subfigure}
\begin{subfigure}{0.35\textwidth}
    \centering
    \includegraphics[width=\textwidth]{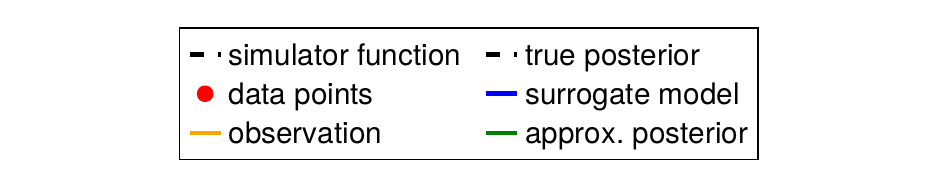}
\end{subfigure}

\caption{Comparison of BO (left) and BOSIP (right) on trivial example with uniform prior $p(x)$, linear simulator $y = x$, normal likelihood $p(z|y) = \mathcal{N}(y,0.2)$, and observation $z_o=0$. The MAP solution is $\hat{x}=0$ and the full Bayesian posterior is $p(x|z_o)=N(0,0.2)$. The classic BO-based approach focuses data at the mode of the posterior in order to find the maximizer $\hat{x}$, while BOSIP spreads data across the high-probability regions in order to learn the shape of the posterior $p(x|z_o)$.} \label{fig:bo-vs-bosip}
\end{figure*}

Computer simulators are prevalent in modern science, providing a means to study complex physical systems across domains ranging from fluid dynamics \cite{kenway2019effective}, computational biology \cite{lakrisenko2023efficient}, ecology \cite{lyu2018adjoint}, epidemiology \cite{wilk2022impact}, etc. A common task is simulation-based inference (SBI): given observations of a real-world system, one seeks to recover the posterior distribution over the simulator's input parameters that are consistent with those observations. This arises naturally when calibrating the simulator to match observed data, or when seeking to estimate quantities of the physical system which cannot be directly measured. The parameter posterior can be directly evaluated for given parameters by running the simulator and evaluating how well the outputs correspond to the observed data under an observation likelihood model. By repeating this process, the full posterior distribution can be characterized. However, this inference task becomes challenging when the simulator is computationally expensive. In many settings of practical interest, a single forward simulation may require hours of compute \cite{bharti2024cost}, rendering exhaustive posterior exploration infeasible. With a limited simulation budget, one must carefully select which parameter configurations to evaluate.

Bayesian optimization (BO) offers a principled framework for this challenge. Rather than using the expensive simulator directly, one maintains a probabilistic surrogate model that approximates the simulator function. The most common choice is the Gaussian process (GP) model. The surrogate allows for a cheap posterior approximation and a principled way to guide further simulator queries through the maximization of an acquisition function, which is evaluated based on the surrogate's predictions. This approach to simulator inference has received significant attention \cite{jarvenpaa2021parallel,surer2026batch,villani2024adaptive,wang2018adaptive,helin2022introduction,reiser2025uncertainty} and was recently unified by the authors under the \emph{Bayesian Optimization for Simulator Inverse Problems} (BOSIP) framework \cite{soldat2026bayesian}.

To address the task of posterior characterization, the classic BO procedure must be substantially adapted. Classical BO aims to merely optimize some objective (e.g., find \emph{the} parameters most consistent with the observed data), and to this end employs optimization-focused acquisition functions like the Expected Improvement. In contrast, BOSIP addresses the active learning problem of estimating the full parameter posterior distribution. This fundamentally different objective requires propagating the surrogate's predictive uncertainty through the likelihood to obtain a notion of the uncertainty in the posterior PDF estimate. The posterior uncertainty is essential for the evaluation of the specialized information-focused acquisition functions, such as Maximum Variance or Expected Integrated Variance \cite{jarvenpaa2021parallel} used in BOSIP. Consequently, BOSIP acquires different data than the classical BO approach. The difference between BO and BOSIP is illustrated in Figure \ref{fig:bo-vs-bosip}, which shows iterations 5 and 10 of both approaches on a trivial toy problem with a linear simulator function, Gaussian likelihood, and a uniform parameter prior. The classical BO approach (left) seeks to maximize the log-posterior probability, which is equivalent to the mean-squared error (MSE) in this case. The surrogate model directly approximates the MSE, and the Expected Improvement acquisition concentrates data around the mode of the posterior. The BOSIP approach (right) instead models the simulator outputs, and the surrogate of the simulator is then used to compute an approximation of the full posterior distribution. The MaxVar acquisition function selects points that maximize the uncertainty in the posterior PDF estimate. Note that the uncertainty in the posterior PDF is distinct from the surrogate's uncertainty in the prediction of the simulator outputs, as it is distorted by propagating the surrogate's predictive distribution through the likelihood and multiplying by the prior. This approach spreads the data more evenly across the high-probability regions in order to identify the exact shape of the posterior.

Meanwhile, simulators based on differential equations increasingly offer access to gradients through automatic differentiation frameworks \cite{chen2018neural,rackauckas2020universal,ma2021comparison,johnson2012notes,jax2018github,Zygote.jl-2018}. Crucially, the cost of reverse-mode gradient computation only scales with the output dimensionality, providing cheap gradient information in high-dimensional parameter settings. In contrast, the cost of forward-mode differentiation scales with input dimensionality, making it suitable when the number of outputs exceeds parameter dimensionality. Despite the growing availability of simulator gradients, they remain largely untapped in the context of simulation-based inference. This represents an unrealized potential, as with expensive simulators we seek to maximize the information gathered from each simulation to remain sample-efficient. There exist well-established techniques for incorporating derivative information directly into Gaussian processes \cite{wu2017bayesian,williams2006gaussian,solak2002derivative}. This yields a gradient-enhanced GP surrogate, which propagates gradient information into the predictions of the functional values themselves, improving predictive performance.

In this work, we bring gradient-enhanced GPs into Bayesian simulation-based inference. Our contributions are as follows:
We describe how gradients can be incorporated into GP-based surrogate models within the BOSIP inference loop, and show that this substantially accelerates posterior convergence and improves approximation quality within a limited simulation budget.
We validate our approach across seven benchmarks (including analytical toy problems and PDE-governed simulators), achieving significant and consistent performance improvements over gradient-free baselines.
We show that in problems with higher parameter dimensionality than output dimensionality, reverse-mode differentiation maintains performance benefits when accounting for the additional computational cost. In contrast, in problems with high output dimensions, the inference speed-up from forward-mode differentiation does not outweigh the added computational costs. Therefore, the benefits of gradient-enhanced surrogates are primarily realized in problems with lower output dimensions, where the reverse-mode is efficient.
Finally, we demonstrate that the benefits of reverse-mode gradients increase in higher-dimensional parameter spaces, representing a promising avenue for future work.

The paper is structured as follows: In section \ref{sec:bosip}, we define the simulator inference problem and present the BOSIP approach. In section \ref{sec:grad-bosip}, we show how gradient information can be cleanly incorporated into the method. In section \ref{sec:experiments}, we demonstrate the performance benefits of exploiting gradients on several benchmarks. Lastly, in section \ref{sec:discussion}, we discuss practical implications of our findings and limitations of the proposed approach.

\section{Simulator Inference via Bayesian Optimization} \label{sec:bosip}

This section recaps the key theoretical and methodological framework of the \emph{Bayesian Optimization for Simulator Inverse Problems} (BOSIP) method proposed by the authors in \cite{soldat2026bayesian}. We refer the reader to that work for a comprehensive description of the method, including a detailed discussion of the design choices and a unifying perspective on related work.

\subsection{Problem Formulation}

We consider a real-world experiment with unknown quantities $x_\text{true}$ and $y_\text{true}$, corresponding to the input and the output of the simulator, respectively. From this experiment, we observe a noisy measurement $z_o \sim p(z|y_\text{true})$, where the observation model $p(z|y)$ describes the uncertainty in the measurement. We have access to an \emph{expensive} high-fidelity simulator that models the underlying physical process, which can be queried with input parameters $x$ to produce outputs $y$. Our goal is to infer the true parameters $x_\text{true}$ that are consistent with the observed data $z_o$. Note that in general, the observation could depend on both quantities, $z \sim p(z|x,y)$, but we omit the dependence on $x$ for simplicity. The methodology can be easily extended to this more general case.

We formalize the problem as Bayesian inference, where we seek to identify the posterior distribution $p(x|z_o)$ of the simulator input parameters given the observed data. Using Bayes' theorem, the posterior is expressed as
\begin{equation}
    p(x|z_o) = \frac{p(z_o|x) p(x)}{p(z_o)},
\end{equation}
where $p(z_o|x)$ is the likelihood, $p(x)$ is the parameter prior, and $p(z_o)$ is the evidence. The evidence is unknown but constant with respect to $x$, so we simply focus on learning the unnormalized posterior $p(z_o|x) p(x)x$,
while the normalization constant
\begin{equation}
    p(z_o) = \int p(z_o|x) p(x) \, dx
\end{equation}
can be recovered post-hoc if needed. The prior encodes our beliefs about the parameters before observing the data, and is defined in a closed form based on domain knowledge.

This leaves the likelihood $p(z_o|x)$ as the key quantity to characterize. The likelihood can be rewritten in terms of the simulator outputs $y$ as
\begin{equation}
    p(z_o|x) = p(z_o|y=f(x)).
\end{equation}
Notice that the likelihood can be directly evaluated for any parameters $x$ using the simulator $f$ and the closed-form observation model $p(z_o|y)$. The primary challenge is that characterizing the full posterior requires many likelihood evaluations, which is infeasible under a limited simulation budget. This motivates two key ideas of the BOSIP approach: (1) use a cheap probabilistic surrogate model to approximate the simulator and thus obtain a cheap posterior approximation, and (2) use Bayesian optimization in order to sequentially select parameters that maximize information gain about the posterior, enabling sample-efficient learning.

\subsection{Surrogate-Based Posterior Approximation}

Our ultimate goal is to learn the probability density function of the parameter posterior $p(x|z_o)$. To this end, we define the auxiliary variable
\begin{equation}
    s(x) = p(z_o|y=f(x)) p(x),
\end{equation}
representing the unnormalized posterior probability, and we pose the function $x \mapsto s$ as our target of inference. We also define a second auxiliary variable
\begin{equation}
    \ell = p(z_o|y=f(x)),
\end{equation}
representing the likelihood probability value, which is the key to the inference as it contains the expensive simulator $f$. To address the expensive simulator cost, BOSIP introduces a surrogate model $\mathcal{M}$ that approximates a \emph{proxy variable}
\begin{equation}
    \delta = (\phi \circ f)(x),
\end{equation}
defined by the proxy mapping $\phi: y \mapsto \delta$, an arbitrary transformation of the simulator outputs. This allows for a flexible choice of a suitable surrogate target. The likelihood estimate is then obtained by applying a complementary likelihood mapping $\psi: \delta \mapsto \ell$.

Considering a dataset $\bm{D}_n=\{(x_i, \delta_i)\}_{i=1}^n$ of previous simulator evaluations, the surrogate model provides the predictive distribution
\begin{equation}
    \hat\delta \sim \mathcal{M}(\delta|x,\bm{D}_n),
\end{equation}
describing the surrogate's estimate of the proxy variable $\delta$ corresponding to the given parameters $x$. We will use the notation $\mathcal{M}_n(\delta|x)$ for brevity. The predictive distribution of the surrogate allows us to estimate the (unnormalized) posterior probability
\begin{equation} \label{eq:post-mean}
    \mathbb{E}_{\delta(x)|\mathcal{M}_n}\left[s(x)\right] = p(x) \mathbb{E}_{\delta(x)|\mathcal{M}_n}\left[\ell(x)\right],
\end{equation}
where the likelihood estimate
\begin{equation}
    \mathbb{E}_{\delta(x)|\mathcal{M}_n}\left[\ell(x)\right] = \int \psi(\delta) \mathcal{M}_n(\delta|x) \,d\delta.
\end{equation}
is obtained by marginalizing over the predictive uncertainty, without requiring additional simulator evaluations. Moreover, we can propagate the surrogate's predictive uncertainty through the likelihood and obtain the variance of the unnormalized posterior
\begin{multline} \label{eq:post-var}
    \mathbb{V}_{\delta(x)|\mathcal{M}_n}\left[s(x)\right] =\\ p(x)^2 \left( \mathbb{E}_{\delta(x)|\mathcal{M}_n}\left[\ell(x)^2\right] - \mathbb{E}_{\delta(x)|\mathcal{M}_n}\left[\ell(x)\right]^2 \right)
\end{multline}
as a measure of the uncertainty in our posterior estimate. This is crucial to the second key part of the BOSIP method: the active data acquisition strategy described in section \ref{sec:bosip-bo}.

\subsection{Gaussian Process Surrogate} \label{sec:bosip-gp}

We model each proxy dimension $\delta^j$ with an independent Gaussian process. The predictive distribution is defined in a closed form as a normal distribution
\begin{equation} \label{eq:gp}
    \mathcal{M}_n(\delta^j|x) = \mathcal{N}(\mu_n(x), \sigma_n^2(x)),
\end{equation}
where the predictive mean $\mu_n(x)$ and variance $\sigma_n^2(x)$ are given by the standard GP regression equations
\begin{equation} \label{eq:gp-mean}
    \mu_n(x) = \mu_0(x) + \bm{k}(x)^T \left(\bm{K} + \sigma_\text{sim}^2 \bm{\mathrm{I}}\right)^{-1} (\bm\delta^j - \bm{m}),
\end{equation}
\begin{equation} \label{eq:gp-var}
    \sigma_n^2(x) = k(x,x) - \bm{k}(x)^T \left(\bm{K} + \sigma_\text{sim}^2 \bm{\mathrm{I}}\right)^{-1} \bm{k}(x).
\end{equation}
The GP model is defined by its prior mean function $\mu_0(x)$ and the kernel $k(x,x')$. The terms which scale with the number of data points $n$ are highlighted in bold: $\bm{K}$ is the $n \times n$ covariance matrix of the training inputs, $\bm{k}(x)$ is the $n$-dimensional covariance vector between the training inputs and the test input $x$, $\bm{\delta}^j$ is the $n$-dimensional vector of training targets, and $\bm{m}$ is the $n$-dimensional vector of prior mean values at the training inputs.

The prior mean function is typically set to a constant zero-mean $\mu_0(x) = 0$ as a default choice, unless prior knowledge about the simulator is available. The kernel function $k(x,x') \in \mathbb{R}$ encodes assumptions about the smoothness and structure of the modeled response surface. Common choices include the Matérn kernels and the squared exponential (RBF) kernel
\begin{equation}
    k(x,x') = \sigma_0^2 \exp\left(-\frac{1}{2} (x-x')^T \Lambda^{-1} (x-x')\right),
\end{equation}
where the diagonal matrix $\Lambda = \text{diag}(\lambda)$ contains the lengthscale parameters $\lambda$ that control the correlation distance, and $\sigma_0^2$ controls the amplitude of the modeled function.

We use the maximum a posteriori (MAP) estimate
\begin{equation}
    \theta^* \in \arg\max_\theta p(\theta|\bm{D}_n)
\end{equation}
for hyperparameter selection, where $\theta$ contains the simulation noise deviation $\sigma_\text{sim}$ and the kernel hyperparameters: lengthscales $\lambda$ and amplitude $\sigma_0$. Note that each proxy dimension has its own set of hyperparameters, but we omit the superscripts $j$ to reduce clutter. The posterior distribution over the hyperparameters can be computed in terms of the marginal data likelihood as
\begin{equation} \label{eq:hyperparameter-posterior}
    p(\theta|\bm{D}_n) \propto p(\bm{\delta}|\bm{x},\theta) p(\theta),
\end{equation}
where $p(\theta)$ is a user-defined hyperparameter prior, and the marginal data likelihood $p(\bm{\delta}|\bm{x},\theta)$ can be computed in closed form:
\begin{equation}
    p(\bm{\delta}|\bm{x},\theta) = \mathcal{N}(\bm{\delta}|\bm{m}, \bm{K} + \sigma_\text{sim}^2 \bm{\mathrm{I}}).
\end{equation}

An advantage of the GP surrogate is that the posterior mean \eqref{eq:post-mean} and variance \eqref{eq:post-var} can be computed in closed form in some cases: for example, when the likelihood is Gaussian and the proxy variable is $\delta = y$.

\subsection{Active Data Acquisition} \label{sec:bosip-bo}

\begin{figure*}[t]
\begin{minipage}{0.48\textwidth}
\begin{algorithm}[H]
\caption{BOSIP} \label{alg:bosip}

\textbf{Input:}
Simulator function $f: x \mapsto y$, proxy mapping $\phi: y \mapsto \delta$, likelihood mapping $\psi: \delta \mapsto \ell$, surrogate model $\mathcal{M}$, acquisition function $\alpha$, prior $p(x)$, initial dataset $\bm{D}_{n_0}$, simulation budget $N$

\textbf{Output:} Approx. posterior $\hat{p}(x|z_o) \approxpropto p(x|z_o)$

\begin{algorithmic}[1]
\State $\mathcal{M}_{n_0} \gets \text{Model}(\bm{D}_{n_0},\mathcal{M})$
\State $\alpha_{n_0} \gets \text{Acq}(\mathcal{M}_{n_0},\psi,p(x))$
\For{$n = n_0+1,...,N$}
\State $x_n \gets \arg\max_x \alpha_{n-1}(x)$ \Comment{Select next params}
\State $\delta_n \gets (\phi \circ f)(x_n)$ \Comment{Evaluate simulator}
\State $\bm{D}_n \gets \bm{D}_{n-1} \cup \{(x_n,\delta_n)\}$ \Comment{Extend dataset}
\State $\mathcal{M}_n \gets \text{Model}(\bm{D}_n,\mathcal{M})$ \Comment{Update surrogate}
\State $\alpha_n \gets \text{Acq}(\mathcal{M}_n,\psi,p(x))$ \Comment{Construct acq.func.}
\EndFor
\State $\hat{p}(x|z_o) = \mathbb{E}_{\delta(x)|\mathcal{M}_N}\left[\psi(\delta)p(x)\right]$ \Comment{eq. \eqref{eq:post-mean}}
\end{algorithmic}

\end{algorithm}
\end{minipage}
\hfill
\begin{minipage}{0.48\textwidth}
\begin{algorithm}[H]
\caption{BOSIP with Gradients} \label{alg:grad-bosip}

\textbf{Input:}
Simulator with gradients $\tilde{f}: x \mapsto (y,\nabla_x f)$, proxy mapping $\phi: y \mapsto \delta$, likelihood mapping $\psi: \delta \mapsto \ell$, surrogate model $\mathcal{M}$, acquisition function $\alpha$, prior $p(x)$, initial dataset $\bm{\tilde{D}}_{n_0}$, simulation budget $N$

\textbf{Output:} Approx. posterior $\hat{p}(x|z_o) \approxpropto p(x|z_o)$

\begin{algorithmic}[1]
\State $\mathcal{\tilde{M}}_{n_0} \gets \text{Model}(\bm{\tilde{D}}_{n_0},\mathcal{\tilde{M}})$
\State $\alpha_{n_0} \gets \text{Acq}(\mathcal{\tilde{M}}_{n_0},\psi,p(x))$
\For{$n = n_0+1,...,N$}
\State $x_n \gets \arg\max_x \alpha_{n-1}(x)$
\State $y_n, \frac{\partial y}{\partial x_n} \gets f(x_n)$ \Comment{Adjoint gradient}
\State $\nabla_x \delta(x_n) \gets \frac{\partial \delta}{\partial y_n} \cdot \frac{\partial y}{\partial x_n}$ \Comment{Chain rule}
\State $\bm{\tilde{D}}_n \gets \bm{\tilde{D}}_{n-1} \cup \{(x_n,\delta_n,\nabla_x \delta(x_n))\}$
\State $\mathcal{\tilde{M}}_n \gets \text{Model}(\bm{\tilde{D}}_n,\mathcal{\tilde{M}})$ \Comment{Augmented model}
\State $\alpha_n \gets \text{Acq}(\mathcal{\tilde{M}}_n,\psi,p(x))$ \Comment{Adapted acq.func.}
\EndFor
\State $\hat{p}(x|z_o) = \mathbb{E}_{\delta(x)|\mathcal{\tilde{M}}_N}\left[\psi(\delta)p(x)\right]$ \Comment{eq. \eqref{eq:post-mean}}
\end{algorithmic}

\end{algorithm}
\end{minipage}
\end{figure*}

Up to this point, we have described how to obtain a cheap posterior approximation in order to avoid expensive simulator evaluations. However, the quality of this approximation depends on the data used to fit the surrogate. With a limited simulation budget, it is crucial to strategically select which parameter configurations to evaluate in order to maximize the information gained about the posterior distribution. This is where Bayesian optimization comes into play.

To address the active data acquisition challenge, Bayesian optimization sequentially selects the most informative simulation parameters. The BO loop cycles through three main steps: (1) fit the surrogate model $\mathcal{M}_n$ based on the current data $\bm{D}_n$, (2) select the next evaluation point
\begin{equation}
    x_{n+1} = \arg\max_x \alpha_n(x),
\end{equation}
by maximizing the acquisition function $\alpha_n$, and (3) evaluate the simulator at the selected parameters to obtain the augmented dataset $\bm{D}_{n+1} = \bm{D}_n \cup \{(x_{n+1}, \delta_{n+1})\}$. This cycle continues until the simulation budget is exhausted. The whole procedure is summarized in Algorithm \ref{alg:bosip} borrowed from \cite{soldat2026bayesian}.

The key to sample-efficient inference is the acquisition function, which balances the exploration of regions where the predictive uncertainty is high, and exploitation of regions of large estimated posterior probability. The acquisition function is evaluated using the surrogate model, and thus can be efficiently maximized without running the simulator. The default choice in BOSIP is the MaxVar acquisition function, which simply selects the point that maximizes the posterior variance
\begin{equation}
    x_{n+1} = \arg\max_x \mathbb{V}_{\delta(x)|\mathcal{M}_n}\left[s(x)\right],
\end{equation}
given by \eqref{eq:post-var}. Since the predictive distribution is transformed through the likelihood mapping and multiplied by the prior, the posterior variance naturally balances exploration and exploitation. The MaxVar acquisition performs well in practice and can be computed in closed form in some cases.

\section{Incorporating Gradient Information} \label{sec:grad-bosip}

To facilitate the integration of gradient information, we adopt the gradient-enhanced GP formulation from \cite{wu2017bayesian}, which extends the standard GP to simultaneously condition on function values and their partial derivatives. Each proxy dimension $\delta^j$ is now represented by a multidimensional GP, where both the function value $\delta^j$ and its $d$-dimensional gradient $\nabla_x \delta^j$ are modeled jointly. The gradient-enhanced GP is defined by its augmented mean and kernel functions
\begin{equation}
\tilde{\mu}_0(x) = \begin{pmatrix} \mu_0(x) \\ \nabla_x \mu_0(x) \end{pmatrix}, \quad
\tilde{K}(x, x') = \begin{pmatrix}
K(x, x') & J(x, x')\\
J(x', x)^T & H(x, x')
\end{pmatrix},
\end{equation}
where $K(x, x')$ is the standard scalar kernel function,
\begin{equation}
    J(x, x') = \left(\frac{\partial K(x, x')}{\partial x'_1}, \ldots, \frac{\partial K(x, x')}{\partial x'_d}\right)^T
\end{equation}
is the $d$-dimensional Jacobian vector of kernel partial derivatives, and $H(x, x')$ is the $d \times d$ Hessian matrix of second partial derivatives. The new predictive distribution remains analogous to \eqref{eq:gp-mean} and \eqref{eq:gp-var}, except we substitute the augmented mean and kernel functions and the augmented $n(1+d)$-dimensional target vector
\begin{eqnarray}
    \bm{\tilde\delta}^j = (\delta^j_1, \nabla_x \delta^j(x_1), ..., \delta^j_n, \nabla_x \delta^j(x_n))^T,
\end{eqnarray}
now including both the function values and gradients from previous simulations. The full kernel matrix $\tilde{\bm{K}}$ becomes an $n(1+d) \times n(1+d)$ matrix, where the block corresponding to data points $i$ and $j$ is given by $\tilde{K}(x_i, x_j)$. This way, the kernel matrix includes the covariances between function values, between gradients, and the cross-covariances between function values and gradients.

We also introduce the augmented dataset
\begin{equation}
    \bm{\tilde{D}}_n = \{(x_i, \delta_i, \nabla_x \delta(x_i))\}_{i=1}^n,
\end{equation}
where in a slight abuse of notation $\nabla_x \delta(x_i)$ represents the full Jacobian matrix of the complete proxy variable $\delta$ w.r.t. the parameters $x$. The Jacobian is calculated using the chain rule
\begin{equation}
    \nabla_x \delta(x_i) = \frac{\partial \delta}{\partial y_i} \cdot \frac{\partial y}{\partial x_i} ,
\end{equation}
where the first term is the Jacobian of the proxy mapping $\phi: y \mapsto \delta$, computed analytically given the closed-form definition of $\phi$, and the second term is the Jacobian of the simulator $f: x \mapsto y$, which is obtained from the simulator via forward or reverse-mode automatic differentation.

The gradient augmentation introduces no additional kernel hyperparameters, as the augmented kernel is fully determined by the original kernel (and its derivatives). However, we introduce additional noise hyperparameters: for each proxy dimension $j$, we now have a $d$-dimensional vector of gradient noise deviations $\sigma_\text{grad}^j$, in addition to the original scalar simulation noise deviation $\sigma_\text{sim}^j$. These gradient noise deviations capture the measurement uncertainty in each partial derivative, and are added to the diagonal of the kernel matrix in \eqref{eq:gp-mean} and \eqref{eq:gp-var} in place of $\sigma_\text{sim}$ for the corresponding gradient observations. The hyperparameters are still selected by maximizing the hyperparameter posterior \eqref{eq:hyperparameter-posterior}.

In addition to the augmented surrogate model, the acquisition function must be adapted as well to account for the new predictive distribution. In case of the MaxVar acquisition, the posterior variance in equation \eqref{eq:post-var} is now computed using the marginal predictive distribution $\mathcal{\tilde{M}}_n(\delta|x)$ of the proxy variable, which is easily extracted from the full predictive distribution $\mathcal{\tilde{M}}_n(\delta,\nabla_x \delta|x)$ due to the Gaussianity of the model. In case of the more complex acquisition functions, such as the ones discussed in \cite{wu2017bayesian} and \cite{soldat2026bayesian}, the adaptation is more involved, as we need to marginalize over the gradients. We leave the analysis of more complex acquisition functions for future work, as we only benchmark using the MaxVar acquisition in this study.

Similarly, the posterior estimate \eqref{eq:post-mean} is obtained simply by substituting the marginal predictive distribution $\mathcal{\tilde{M}}_n(\delta|x)$ of the proxy variable. However, the gradient observations still improve the accuracy of the posterior estimate, as the gradient information is propagated through the augmented covariance structure of the model into the predictions of the proxy values. The whole adapted procedure is summarized in Algorithm \ref{alg:grad-bosip}.

\section{Experiments} \label{sec:experiments}

\begin{figure*}[t]
\centering
\includegraphics[width=\textwidth]{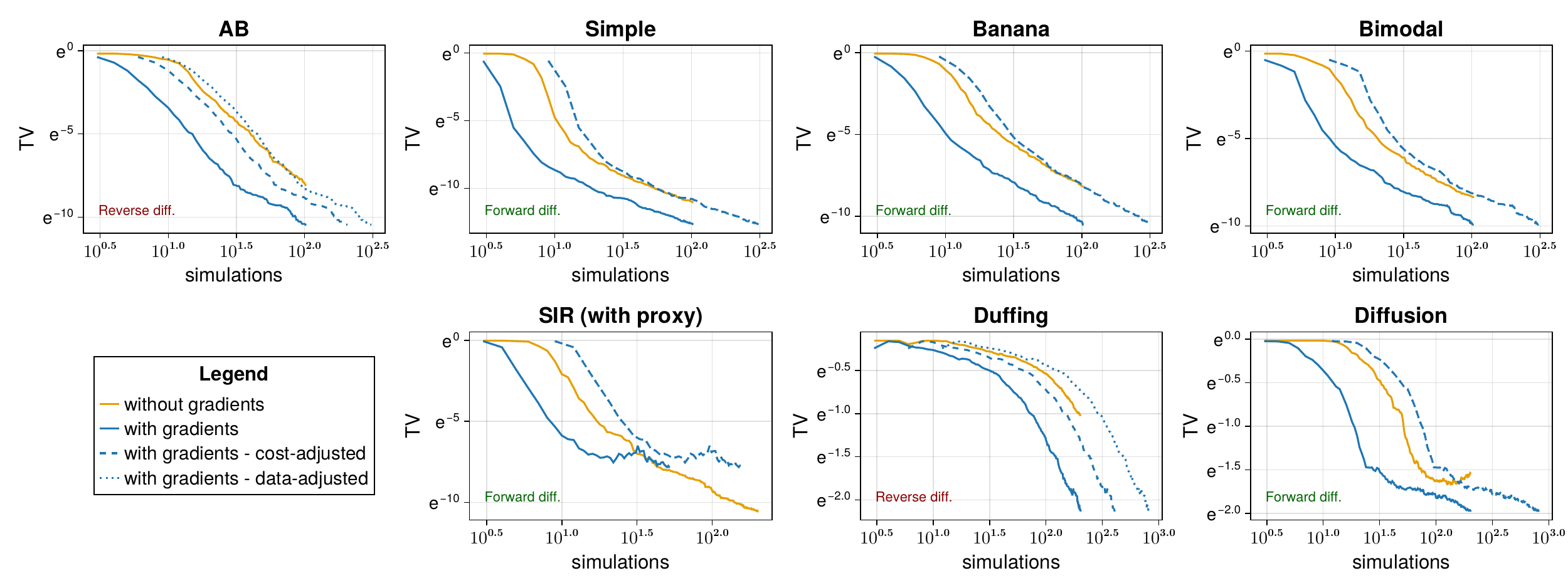}
\caption{Comparison of BOSIP with and without gradients across seven benchmarks. The solid lines show median scores across 20 runs. The dashed lines show the scores adjusted for the additional computational cost of the gradients. The dotted lines show the score adjusted for the total information obtained from the simulator. Note that the cost-adjusted and data-adjusted scores are equal in case of forward-mode gradients. Importantly, when reverse-mode is used, the performance of the gradient-enabled BOSIP is better than the gradient-free BOSIP even when accounting for the additional computational cost.
}
\label{fig:results}
\end{figure*}

\begin{figure*}[t]
\centering
\includegraphics[width=\textwidth]{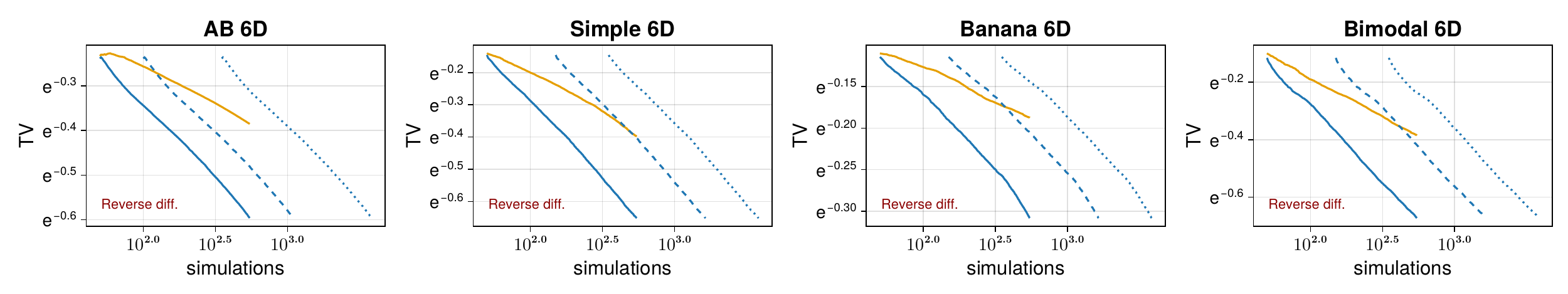}
\caption{Performance on higher-dimensional benchmarks. The performance benefits of exploiting reverse-mode gradients are larger in higher dimensions. For the same cost of a reverse-mode gradient pass, we obtain more gradient information, improving inference efficiency further.}
\label{fig:multidim}
\end{figure*}

\begin{figure*}[t]
\centering
\includegraphics[width=\textwidth]{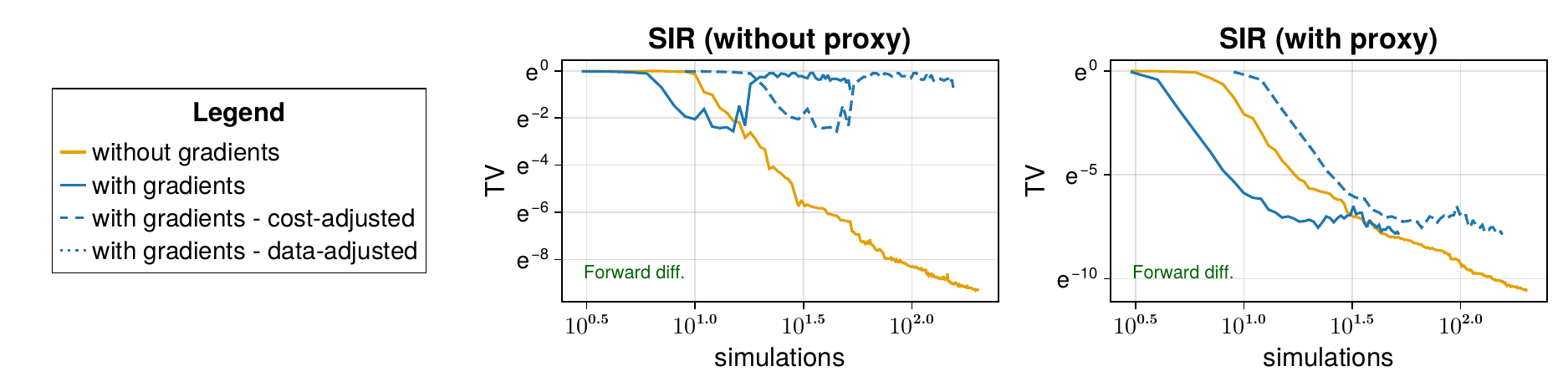}
\caption{Impact of suitable proxy variable design on the SIR benchmark. The response surface of the SIR simulator exhibits sharp ridges, where the gradients attain huge values, causing numerical issues in the gradient-enhanced surrogate. The performance is greatly improved by using a proxy variable that smooths the response surface, significantly mitigating the numerical issues. The gradient-enhanced runs are truncated to 50 simulations, as the numerical issues impede further convergence.}
\label{fig:sir}
\end{figure*}

We validate the gradient-enhanced BOSIP approach on the seven benchmark problems used in \cite{soldat2026bayesian}, some of which are originally adapted from \cite{lueckmann2021benchmarking}. The benchmarks include 4 analytical toy problems with varying complexity, as well as 3 simulators governed by partial differential equations that represent more realistic inference scenarios, including 2D and 3D problems. In addition, we extend the analytical problems to 6D parameters to showcase the increased benefits of exploiting gradients in a higher-dimensional setting. The gradient-enhanced BOSIP is compared against the standard gradient-free approach from \cite{soldat2026bayesian}. We evaluate the performance in terms of the quality of the posterior approximation obtained within a limited simulation budget. Each experiment is repeated with 20 random sets of 3 initial data points each, and we report the median performance across these runs. The simulation budgets are set to 100 simulations for the analytical problems and 200 simulations for the PDE-based simulators. The 6D problems are evaluated with a budget of 500 simulations, which is not sufficient for convergence, but still allows to demonstrate the behavior in higher dimensions. The 6D experiments were performed without kernel hyperparameter optimization.

For detailed descriptions of the 2D and 3D benchmarks, including visualizations of the true posteriors, see \cite{soldat2026bayesian}. The new 6D benchmarks are constructed by extending the 2D analytical problems as follows. Given the original simulator $f_{2\text{D}}(x_{1:2}) \mapsto y$, we construct the 6D simulator as
\begin{equation}
    f_{6\text{D}}(x_{1:6}) = \frac{1}{3} \left( f_{2\text{D}}(x_{1:2}) + f_{2\text{D}}(x_{3:4}) + f_{2\text{D}}(x_{5:6}) \right).
\end{equation}
Since the 6D simulator outputs a single averaged value, the likelihood and the observation remain the same as in the original 2D problem. This construction retains a similar posterior structure to the original 2D problems.

The quality of the posterior approximation is measured by the total variation (TV) distance
\begin{equation}
    \text{TV}(\hat{p}, p) = \frac{1}{2} \int |\hat{p}(x|z_o) - p(x|z_o)| \, dx,
\end{equation}
of the \emph{normalized} posterior approximation $\hat{p}(x|z_o)$ obtained from BOSIP to the true posterior $p(x|z_o)$ of each benchmark problem. The TV distance is approximated via Monte Carlo integration
\begin{equation}
    \text{TV}(\hat{p}, p) \approx \frac{1}{2k} \sum_{i=1}^k \frac{1}{p(x_i)} |\hat{p}(x_i|z_o) - p(x_i|z_o)|,
\end{equation}
by sampling $x_1,...,x_k \sim p(x)$ from the parameter prior. The missing normalization constant $\hat{p}(z_o)$ is estimated via Monte Carlo integration as well.

Figure \ref{fig:results} shows the results across the seven benchmarks. We report the TV distance as a function of the number of simulator evaluations. We observe significant performance benefits from using gradient information, with the gradient-enhanced BOSIP consistently outperforming the gradient-free baseline in terms of posterior approximation quality. For the gradient-enhanced approach, we additionally report the TV distance adjusted for the added computational cost of the gradients. For each problem, we select the most efficient differentiation mode: reverse-mode for problems where the proxy variable dimensionality is lower than the number of parameters ($\delta_{\text{dim}} < x_{\text{dim}}$), and forward-mode for problems where proxy dimensionality exceeds parameter dimensionality ($\delta_{\text{dim}} \geq x_{\text{dim}}$). The cost-adjusted score incorporates the cost of computing gradients: for reverse-mode problems we multiply the simulation count by $(1 + \delta_{\text{dim}})$, and for forward-mode problems we multiply by $(1 + x_{\text{dim}})$, assuming each gradient pass is roughly equivalent to the cost of one simulation. In practice, the exact cost of gradient computation varies, but it remains a small constant overhead \cite{griewank2008evaluating,baydin2018automatic}, making this assumption reasonable for the purposes of this study. This cost adjustment allows for a fair comparison at approximately equivalent total computational cost. In the reverse-mode problems, the performance benefits are maintained even after cost adjustment, confirming that the improved inference efficiency justifies the additional cost of computing gradients. In contrast, in the forward-mode problems, the cost-adjusted performance is worse than the gradient-free approach, indicating the contrary.

We also report the data-adjusted score, which normalizes the achieved performance with respect to the total number of scalar values obtained from the simulator: output values and gradient components. Each simulation additionally provides $x_{\text{dim}} \cdot \delta_{\text{dim}}$ gradient values. Therefore, the data-adjusted score is obtained by multiplying the simulation cost by $(1 + x_{\text{dim}})$. This metric shows performance relative to the total information extracted from the simulator. The data-adjusted plots provide an interesting perspective on the efficiency of the methods. The gradient-enhanced approach is always less efficient with respect to the total amount of data collected. In case of forward-mode differentiation, the added cost of the gradients is proportional to the amount of additional data as both scale with $(1 + x_{\text{dim}})$, thus resulting in overall worse performance with respect to the computational cost. In contrast, in case of reverse-mode differentiation, the added cost of the gradients is only $(1 + \delta_{\text{dim}})$, where $\delta_{\text{dim}} < x_{\text{dim}}$, meaning we obtain cheaper data. This cost advantage outweights the lower data efficiency, resulting in the improved cost-adjusted performance compared to the gradient-free approach.

Figure \ref{fig:multidim} demonstrates behavior in higher dimensions. While not fully converged, the results show substantially increased speed of convergence. This is not surprising, as reverse-mode differentiation becomes increasingly advantageous in higher-dimensional parameter spaces: for the same cost of each reverse-mode pass, we obtain gradient information across all input dimensions, effectively gathering even cheaper data. This highlights the substantial potential of exploiting gradients for accelerated inference in higher-dimensional problems.

Finally, Figure \ref{fig:sir} showcases the importance of suitable proxy variable design. The SIR benchmark exhibits sharp ridges in the response surface, where the gradients attain huge values. This can hinder the inference efficiency of the BOSIP approach, and the numerical issues are further exacerbated in the gradient-enhanced surrogate. By using a proxy variable that smooths the response surface as in \cite{soldat2026bayesian}, we mitigate the numerical issues, allowing the gradient-enhanced approach to achieve satisfactory convergence within the limited simulation budget. To further improve the numerical stability, one may consider a more careful proxy variable design, apply gradient clipping, or manually increase the gradient noise deviation hyperparameters to ensure a well-conditioned kernel matrix. However, the latter two strategies may come at the cost of reduced predictive accuracy.

\section{Discussion} \label{sec:discussion}

The results demonstrate that gradient-enhanced surrogates can significantly accelerate posterior convergence and improve approximation quality across a range of benchmarks, including both analytical problems and realistic problems. However, the practical benefits depend critically on the computational cost of obtaining the gradients. When accounting for the added computational expense, only reverse-mode (adjoint) differentiation provides real performance improvements over gradient-free baselines, whereas forward-mode differentiation is too expensive to justify the performance gains. In such cases, it is more efficient to simply query additional simulations without gradients rather than incur the computational overhead.

These findings lead to a clear and actionable conclusion: incorporating gradient information into BOSIP is worthwhile primarily in problems where parameter dimensionality exceeds the dimensionality of the proxy variable (the model target). Under these conditions, reverse-mode differentiation is computationally efficient and provides gradient information at low computational cost relative to the alternative of running additional simulations. This advantage becomes even more pronounced in higher-dimensional parameter spaces, where reverse-mode differentiation provides gradient information across all input dimensions at the same cost.

A persisting challenge is the scalability of the Gaussian process surrogate, which becomes computationally demanding as the number of data points increases. Since the benefits of gradient-enhanced surrogates are most pronounced in high-dimensional problems that typically require larger simulation budgets to converge, future work should focus on improving GP scalability or exploring alternative surrogate models that can effectively leverage gradient information while handling larger datasets.

\section*{Acknowledgment}

Part of this work has been carried out within the framework of the EUROfusion Consortium, funded by the European Union via the Euratom Research and Training Programme (Grant Agreement No 101052200 - EUROfusion). Views and opinions expressed are however those of the author(s) only and do not necessarily reflect those of the European Union or the European Commission. Neither the European Union nor the European Commission can be held responsible for them.

The access to the computational infrastructure of the OP VVV funded project CZ.02.1.01/0.0/0.0/16\_019/0000765 ``Research Center for Informatics'' is also gratefully acknowledged.

\bibliography{references.bib}

\end{document}